\documentclass[10pt,twocolumn,letterpaper]{article}

\usepackage[pagenumbers]{wacv} 

\usepackage{graphicx}
\usepackage{amsmath}
\usepackage{amssymb}
\usepackage{pifont}
\newcommand{\cmark}{\color{green}\ding{51}}%
\newcommand{\xmark}{\color{red}\ding{55}}%
\usepackage{booktabs}
\usepackage{microtype}

\usepackage[pagebackref,breaklinks,colorlinks]{hyperref}

\usepackage[capitalize]{cleveref}
\crefname{section}{Sec.}{Secs.}
\Crefname{section}{Section}{Sections}
\Crefname{table}{Table}{Tables}
\crefname{table}{Tab.}{Tabs.}

\usepackage{xspace}
\newcommand{\webAppName}{\textit{PupilSense}\xspace}
\newcommand{\appName}{\textit{ChameleonView}\xspace}
\newcommand{\publicDataParticipantCount}{51\xspace}

\begin{document}

\title{\webAppName: A Novel Application for Webcam-Based Pupil Diameter Estimation}

\author{Vijul Shah$^{2}$
\and
Ko Watanabe$^{1, 2}$
\and
Brian B. Moser$^{1, 2}$
\and
Andreas Dengel$^{1, 2}$
\and
\\
$^1$ German Research Center for Artificial Intelligence (DFKI), Germany\\
$^2$ RPTU Kaiserslautern-Landau, Germany\\
\textit{first.second@dfki.de}
}
\maketitle

\begin{abstract}
    Measuring pupil diameter is vital for gaining insights into physiological and psychological states — traditionally captured by expensive, specialized equipment like Tobii eye-trackers and Pupillabs glasses.
This paper presents a novel application that enables pupil diameter estimation using standard webcams, making the process accessible in everyday environments without specialized equipment.
Our app estimates pupil diameters from videos and offers detailed analysis, including class activation maps, graphs of predicted left and right pupil diameters, and eye aspect ratios during blinks. 
This tool expands the accessibility of pupil diameter measurement, particularly in everyday settings, benefiting fields like human behavior research and healthcare.
Additionally, we present a new open source dataset for pupil diameter estimation using webcam images containing cropped eye images and corresponding pupil diameter measurements.
\end{abstract}

\section{Introduction}
The cognitive state of humans is closely linked to features observable through their eyes. Fortunately, the accessibility of eye monitoring in everyday life is rapidly increasing, exemplified by recent advancements such as Apple's incorporation of camera-based eye tracking features~\cite{apple2024accessibility, greinacher2020accuracy}. However, research in this domain primarily targets blink detection~\cite{hong2024robust} and gaze estimation~\cite{o2023toward, yun2022haze}, employing various methodologies, including the use of biomarkers~\cite{liu2022noncontact}, infrared spectrum reflected from the eyes~\cite{fathi2015camera}, or image-based techniques~\cite{hisadome2024rotation}.
In comparison, fewer explore pupil diameter estimation~\cite{sari2016pupil, caya2022development}, which also plays an undeniably crucial role in determining various physiological and psychological states. This oversight highlights a critical gap in the field, underscoring the need for more comprehensive approaches to fully leverage eye monitoring for cognitive state analysis for many reasons.

Previous studies show that the analysis of pupil diameter serves as an indicator of stress~\cite{pedrotti2014automatic}, attention~\cite{ludtke1998mathematical, van2016pupil}, or cognitive work loads~\cite{kahneman1966pupil, pfleging2016mental, krejtz2018eye}.
In addition, the diameter of the pupil is also closely linked to the activity of the locus coeruleus~\cite{murphy2014pupil, joshi2016relationships}, a brain region critical for managing both short-term and long-term memory functions~\cite{kahneman1966pupil, kucewicz2018pupil}.
Pupil diameter is also used for health check purposes, such as checking pupillary light reflex of patients with intracranial lesions in an intensive care unit~\cite{kotani2021novel}.
Accurate pupil diameter estimation is thus fundamental to enhancing the capabilities of image-based eye tracking.

However, we identify three significant challenges in advancing the field of image-based pupil diameter estimation, which we want to address.
The first challenge lies in collecting ground truth data.
Previous works relied on capturing pupil images and subsequently measuring the diameter in pixels, a time-consuming process that is complicated by increasing participant numbers~\cite{sari2016pupil, caya2022development}.
We overcome this issue by applying a sensor substitution approach using Tobii eye-tracker with Tobii Pro SDK~\cite{TobiiProLab} as a reliable ground truth sensor.
This approach allows for efficient data collection by acquiring ground truth diameter values from the eye-tracker and facial recordings via webcam.

The second challenge concerns data diversity.
Previous studies have varied pupil diameter in participants by altering illumination displays~\cite{shanti2021automated}.
We apply a similar approach, changing the computer display's color during our data collection.
Unlike previous work~\cite{sari2016pupil, shanti2021automated, caya2022development}, we impose fewer constraints, allowing them to choose their seating position and distance from the screen. 
This approach enables us to collect data under more natural, ``in the wild'' conditions, potentially enhancing the empirical validity of findings.

The third challenge is the prediction of pupil diameter itself. 
Previous studies~\cite{gaze360_2019, zhang2020eth, ankur2024appearance} have highlighted that estimating gaze coordinates with a camera involves analyzing images of approximately $60\times36$ pixels~\cite{zhang19mpiigaze}. 
The scale of our images will be smaller for pupil diameter estimation, necessitating analysis at an even finer resolution. 
This makes accurately predicting pupil diameters more complex than gaze estimation and presents a challenging task.

\begin{figure*}
  \centering
  \includegraphics[width=\linewidth]{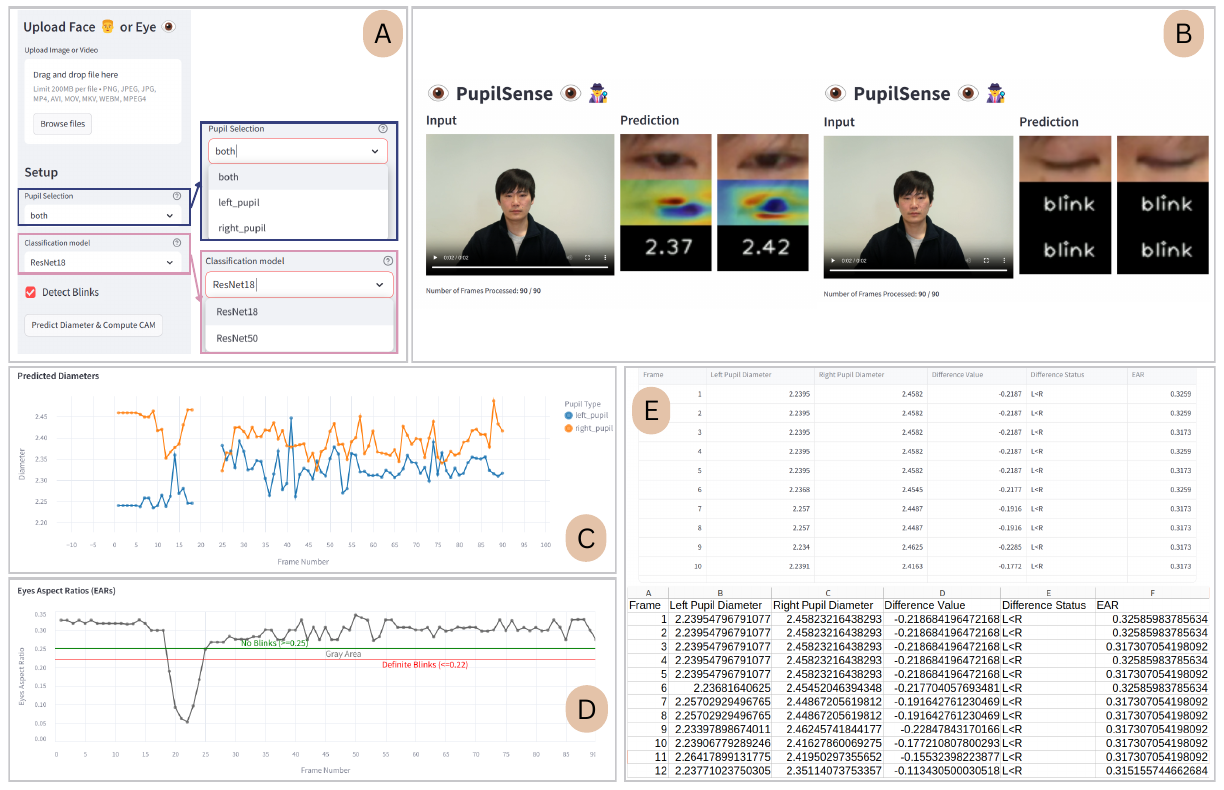}
  \caption{\webAppName: A web app for estimating and analyzing pupil diameters from everyday images and videos.[A]: Options to select either the left or right pupil for analysis (in blue) and to choose the classification models (in pink). [B]: Visualization of the input and output media, including CAM and estimated pupil diameters. [C]: Estimated pupil diameter values for each frame, analyzed by selected pupil type(s). [D]: EAR values for blink detection, with thresholds for acceptance of open eyes (in green) and rejection (in red). [E]: Consolidated data view showing pupil diameter values, EARs, and differences in pupil diameters, with a downloaded CSV file.}
   \label{fig:application}
\end{figure*}

In conclusion, we contribute to the image-based pupil diameter estimation field as follows:

\begin{enumerate} 
\item[C1] We provide an open-source webcam-based pupil diameter estimation dataset.
\item[C2] We propose baseline prediction results using our dataset.
\item[C3] We implement a novel, user-friendly web application for pupil diameter estimation.
\end{enumerate}

\section{Related Work}
This section reviews datasets, methods, and applications for gaze and pupil diameter estimation. We highlight how our dataset addresses the gap in publicly accessible pupil diameter resources, offering the largest collection using RGB webcam images and depth maps.

\autoref{tb:dataset_comparison} compares datasets for gaze and pupil diameter estimation using RGB, RGBD, and IR cameras. While some data were collected in controlled labs, others focused on real-world environments. Most datasets emphasize gaze estimation, highlighting a gap in pupil diameter data, especially in natural settings. Datasets like \textit{Rojas et al.}~\cite{rojas2019pupil}, \textit{Ricciuti et al.}~\cite{ricciuti2020pupil}, and \textit{Caya et al.}~\cite{caya2022development} offer valuable contributions to advancing eye-tracking research. However, they are often limited in scope, providing only numerical pupil diameter values instead of images, or are not publicly accessible. In contrast, our dataset is the largest publicly accessible resource for pupil diameter estimation from RGB images taken from webcam images and additionally computed depth maps, contributing significantly to eye-monitoring research.

\begin{table*}[t!]
  \centering
  \renewcommand{\arraystretch}{1.1}
  \caption{Comparison of related datasets for eye monitoring. While most datasets have gaze coordinates~\cite{funes2014eyediap, zhang19mpiigaze, fischer2018rt, gaze360_2019, lian2019rgbd, zhang2020eth, kothari2020gaze, Kothari2021weaklysupervised, fuhl2021teyed, hou2024multimodal, dembinsky2024eye}, there is a significant gap in pupil diameter informed~\cite{rojas2019pupil, ricciuti2020pupil, caya2022development} datasets.}
  \scalebox{0.82}{
    \begin{tabular}{cccccccccc}
      \hline
        \textbf{Dataset} & Subject & Size & Images & Resolution & Camera & Distance & Gaze Vector & Public & Pupil Diameter \\ 
        \hline
        EyeDiap~\cite{funes2014eyediap} & 16 & 62,500 & \cmark & 1920 x 1080 & RGBD & 80-120 cm & 2D, 3D & \cmark & \xmark \\

        MPIIFaceGaze~\cite{zhang19mpiigaze} & 15 & 213,659 & \cmark & 1280 x 720 & RGB  & varying & 2D, 3D & \cmark & \xmark \\
        
        RT-GENE~\cite{fischer2018rt} & 15 & 122,531 & \cmark & 1920 x 1080 & RGBD & 80-280 cm & 3D & \cmark & \xmark \\
        
        Gaze360~\cite{gaze360_2019} & 238 & 172,000 & \cmark & 4096 x 3382 & RGB & varying & 3D & \cmark & \xmark \\
        
        SHTechGaze+~\cite{lian2019rgbd} & 218 & 165,231 & \cmark & 1920 x 1080 & RGBD & varying & 2D & \cmark & \xmark \\
        
        ETH-XGaze~\cite{zhang2020eth} & 110 & 1,083,492 & \cmark & 6000 x 4000  & RGB & 100 cm & 3D & \cmark & \xmark \\
        
        GW~\cite{kothari2020gaze} & 54 & 5,800,000 & \cmark & 1920 x 1080 & IR & 0.5-3 cm & 2D, 3D & \cmark & \xmark \\
        
        LAEO~\cite{Kothari2021weaklysupervised} & 485 & 800,000 & \cmark & variable & RGB & varying & 3D & \cmark & \xmark \\
        
        \textit{Fuhl et al.}~\cite{fuhl2021teyed}
        & 132 & 20,867,079 & \cmark & variable & IR & 0.5-3 cm & 2D, 3D & \cmark & \xmark \\
        
        \textit{Hou et al.}~\cite{hou2024multimodal} & - & 35,231 & \cmark & 1280 x 720 & RGB & varying & - & \cmark & \xmark \\

        \textit{Dembinsky et al.}~\cite{dembinsky2024eye} & 19 & 648,000 & \xmark & - & - & 67.5 cm & 2D, 3D & \cmark & \xmark \\
        
        \textit{Rojas et al.}~\cite{rojas2019pupil} & 50 & - & \xmark & - & - & 60 cm & 2D, 3D & \cmark & \cmark\\

        \textit{Caya et al.}~\cite{caya2022development} & 16 & - & \cmark & variable & RGB & 10 cm & - & \xmark & \cmark  \\
        
        \textit{Ricciuti et al.}~\cite{ricciuti2020pupil} & 17 & 20,400 & \cmark & 300 x 300 & RGB & - &  - & \xmark & \cmark \\ 
        
        \hline
        
        \textbf{Ours} & \textbf{\publicDataParticipantCount}  & 212,073 & \cmark & 32 x 16  & RGB(D) & varying  & 2D, 3D & \cmark  & \cmark         \\
        
        \hline
      \end{tabular}
  }
  \label{tb:dataset_comparison}
  \caption*{Note: The columns of the above table are: (1) the dataset reference (2) the number of subjects; (3) the size of the dataset (4) images available or not, if not, then it implies that only tabular data are available; (5) the resolution of each image; (6) the type of camera(s), our dataset calculates depth after RGB image recordings and hence represents as RGB(D); (7) the distance to the camera(s); (8) type of gaze vector such as 2D or 3D where ``D'' is a dimention; (9) public dataset or not; and (10) dataset contains pupil diameter or not.}
\end{table*}

Methods for pupil diameter estimation include \textit{PuReST}, developed by \textit{Santini et al.}~\cite{santini2018purest}, which tracks pupils robustly using images from 3 head-mounted devices. Ricciuti and Gambi~\cite{ricciuti2020pupil} employed video processing with the Viola-Jones algorithm for eye cropping and Canny edge detection with Hough transforms for pupil diameter measurement. Similarly, \textit{Caya et al.}~\cite{caya2022development} used preprocessing techniques, including RGB-to-grayscale conversion and the Tiny-YOLO \cite{adarsh2020yolo} algorithm, achieving percent differences of 0.58\% and 0.48\% for the left and right eyes, respectively.

Innovative applications for pupil diameter estimation include \textit{PupilScreen}~\cite{mariakakis2017pupilscreen}, which uses smartphone cameras in a VR-like enclosure for concussion diagnosis, though its fixed pixel-to-millimeter conversion affects accuracy. \textit{Barry et al.}~\cite{barry2023racially} employed a smartphone with an external far-red light attachment, while \textit{Barry et al.}~\cite{barry2022home} used smartphones' NIR and RGB cameras with depth calculations for pupil size estimation. \textit{Strauch et al.}~\cite{strauch2020pupil} demonstrated pupil diameter as a psychophysiological indicator during video gameplay using an SMI-RED 120 eye-tracker. Studies like \textit{Bednarik et al.}~\cite{bednarik2018pupil} linked pupillary responses to expertise in microsurgical training, while \textit{Palinko et al.}~\cite{palinko2012exploring} and \textit{Medathati et al.}~\cite{medathati2020towards} examined cognitive load and state using pupil diameter in driving simulators and real-world settings. Many of these rely on specialized hardware or close-range setups, limiting accessibility. In contrast, \webAppName offers a device-agnostic, hardware-free platform to democratize pupil monitoring, enhancing accessibility and transparency.

\begin{figure*}[t!]
  \centering
  \includegraphics[width=0.8\linewidth]{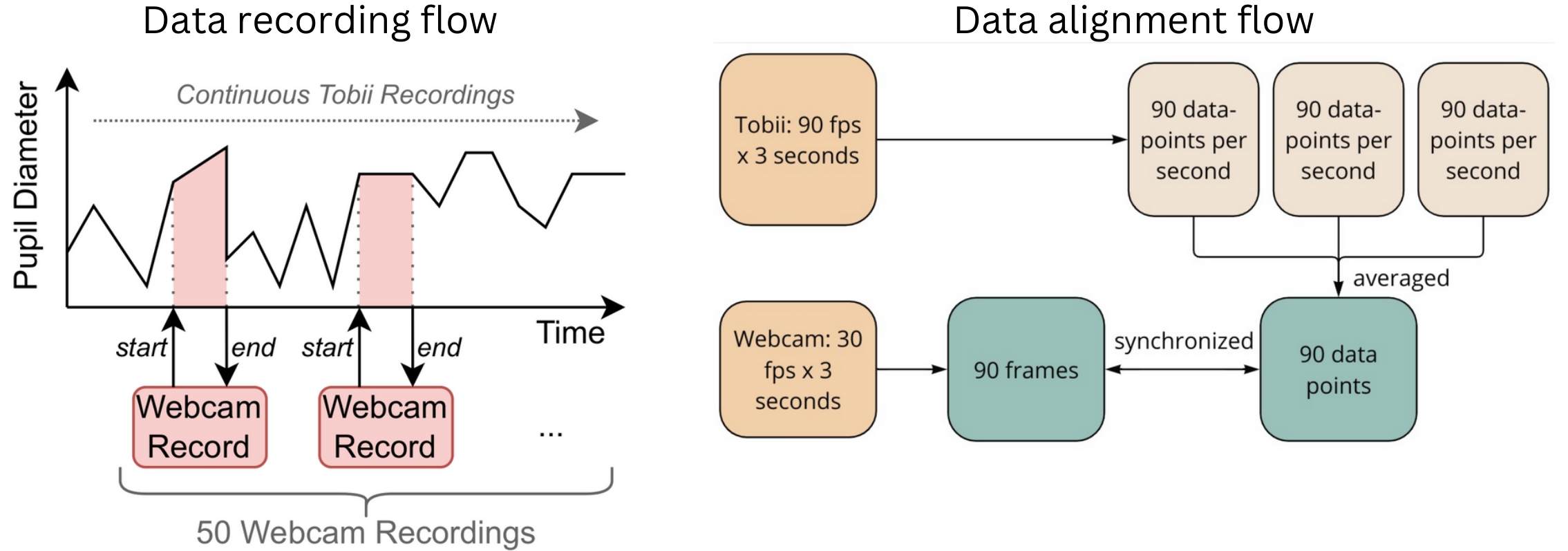}
  \caption{Overview of a data recording and preprocessing (alignment flow). Tobii eye-tracker records pupil diameter, and \appName captures facial recordings using a webcam. Facial recordings start when the participant clicks on the button in the center. The start and end timestamp of the recording is collected in order to synchronize the data with an eye-tracker. To synchronize the 90 frames with the 270 Tobii-captured data points, each metric column is concatenated horizontally across the 90 data points from the three unique timestamps in the Tobii-captured CSV file, followed by computing a row-wise mean.}
  \label{fig:recording_and_alignment_flow}
\end{figure*}

\begin{figure*}[t!]
  \centering
  \includegraphics[width=0.7\linewidth]{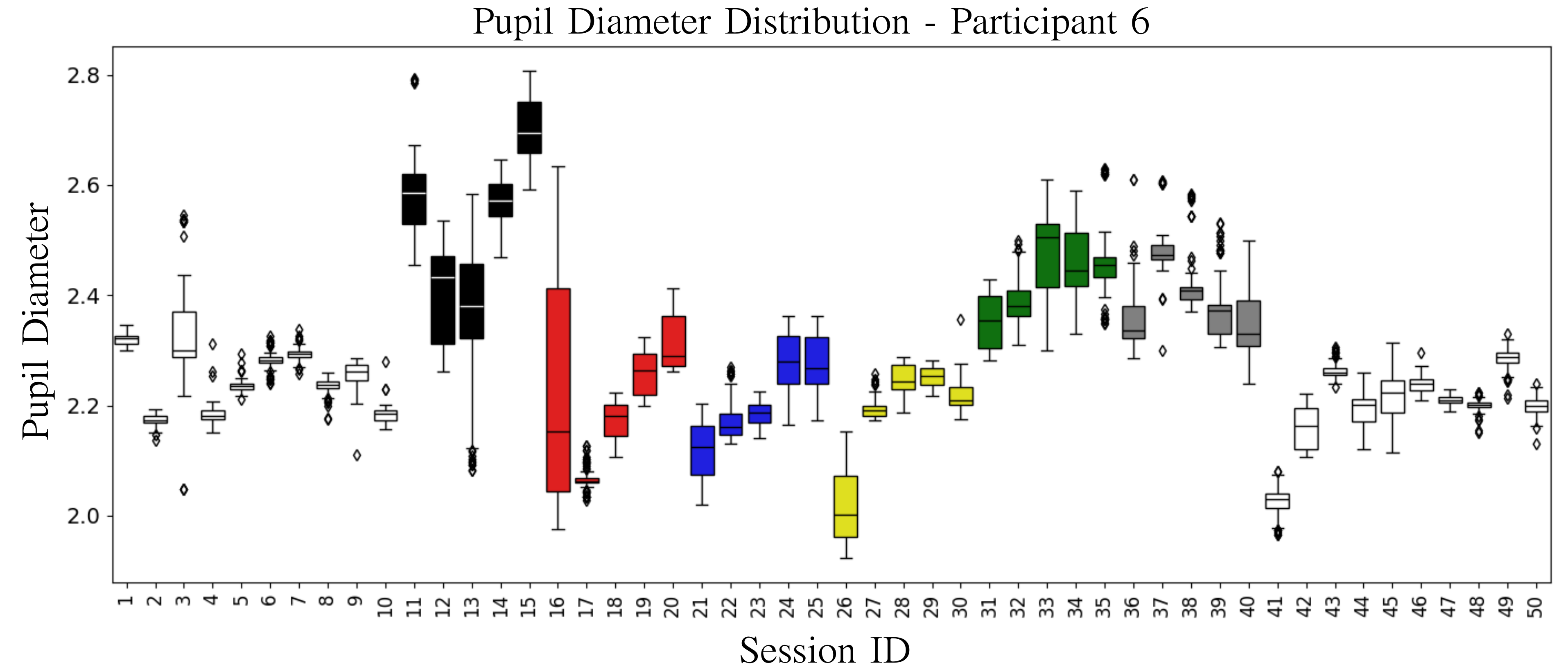}
  \caption{
    Pupil diameter distribution of one participant during the recordings. 
    Different pupil diameter measurements and webcam images were captured during the three-second long sessions (in total, 50 sessions). The colors of the boxes indicate the display color used during the recordings (white, black, red, blue, yellow, green, gray, and white again).
  }
   \label{fig:distribution_pd}
\end{figure*}

\section{Data Collection}
\label{sec:data_collection_app}
We created a novel dataset to address the need for high-quality datasets containing eye images with precise pupil diameter annotations in real-world settings. An overview of the data collection process is shown in \autoref{fig:recording_and_alignment_flow}.

We recruited \publicDataParticipantCount participants (39 males, 11 females, 1 undisclosed, aged 21–44, M=27.58), with consent for the public release of eye-cropped data. Pupil diameter ground truth was recorded at 90 Hz using a Tobii eye-tracker with Tobii Pro Lab software, offering millimeter precision. For video recordings of the face and eyes, we used the built-in webcam of the Microsoft SurfaceStudio 1, which records at a resolution of 1280 × 720 pixels and 30 FPS.

Data was collected using a custom web app, \appName~\footnote{\url{https://chameleon-view.netlify.app/}}, enabling participants to trigger three-second webcam recordings by clicking a central button. As shown in \autoref{fig:recording_and_alignment_flow}, a timestamps file synchronized webcam videos with Tobii data. Each participant completed 50 recordings, with screen background colors varied to capture diverse pupil sizes~\cite{winn1994factors, portengen2023trade}. The first and last 10 recordings used a white background (\#ffffff), while the middle 40 alternated among black (\#000000), red (\#ff0000), blue (\#0000ff), yellow (\#ffff00), green (\#008000), and gray (\#808080). Recordings were conducted in a well-lit room, and the pupil diameter distribution across sessions is visualized in \autoref{fig:distribution_pd}.

\section{Data Preprocessing}
\label{sec:preprocessing}
To build the final dataset, we merged the raw webcam footage with Tobii eye-tracker data through two key phases: alignment of recordings and eye cropping.

\begin{figure*}[t!]
  \centering
  \includegraphics[width=0.85\linewidth]{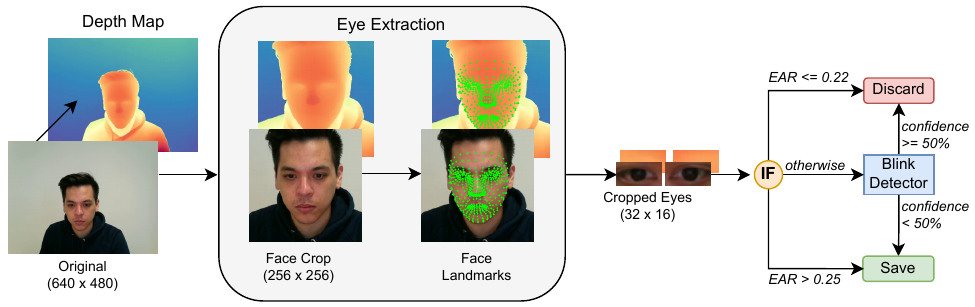}
   \caption{Data preprocessing pipeline to crop the eyes. For face detection and landmark localization, we used Mediapipe to extract the respective cropped eye images (32x16), left and right, separately. We applied a pre-trained DepthAnythingV2 model on the entire image and cropped the depth maps around the eye regions with the help of landmarks detected from Mediapipe. Next, we applied blink detection on the cropped eyes using the Eye Aspect Ratio (EAR) and a pre-trained vision transformer for blink detection. Cropped eye images and the depth maps are then saved based on the EAR threshold and model confidence score.}
   \label{fig:preprocessing_flow}
\end{figure*}

\subsection{Aligning the Recordings}
\label{sec:aligning}
The process of aligning data from the Tobii eye-tracker and webcam recordings is explained as follows:

\begin{enumerate}
    \item \textbf{Data sources:} Tobii eye-tracker captures pupil diameters and gaze positions at 90 data points per second (90Hz) and webcam recordings capture video at 30 frames per second (30fps).
    \item \textbf{Matching timestamps:} Each recording has start and end timestamps, see \autoref{fig:recording_and_alignment_flow} (left). These timestamps are used to extract the corresponding rows from the Tobii eye-tracker data that fall within this time range for synchronization.
    \item \textbf{Frame and data count:} Each recording is 3 seconds long, resulting in: 90 frames from the webcam (3 seconds $\times$ 30fps) and 270 data points from the Tobii eye-tracker (three seconds $\times$ 90Hz).
    \item \textbf{Aligning frames and data:} Concatenate each of the 90 data points from the three unique timestamps and compute a row-wise mean, yielding 90 image frames aligned with 90 data points. And lastly, the first timestamp of the trio is designated as the primary timestamp for that recording, ensuring consistency in data alignment.
\end{enumerate}
This process is repeated for all 50 recording sessions for all 51 participants to ensure uniformity in the dataset.

\subsection{Cropping the Eyes} 
\label{sec:croppings}
After aligning frames, we crop the eye regions using Mediapipe \cite{lugaresi2019mediapipe}, which detects facial landmarks. To ensure consistency despite variations in participants' distance from the webcam, eye regions are cropped to fixed dimensions of 32x16 pixels, preserving the natural shape and scale of the eyes. 

The full process is presented in \autoref{fig:preprocessing_flow}. 
Depth maps are generated for the entire image with face, using the DepthAnythingV2~\cite{yang2024depth} model and cropped based on eye coordinates detected by Mediapipe, allowing us to extract depth information without an RGBD camera.
To remove frames with blinks, we use the Eye Aspect Ratio (EAR) calculated from Mediapipe landmarks. Frames with EAR <= 0.22 are classified as blinks, while EAR > 0.25 indicates open eyes. A Vision Transformer (ViT) model \cite{kaggle_closed_eye_detection} enhances classification for ambiguous cases (EAR > 0.22 and <= 0.25). Frames with blinks are discarded. Given blink durations of 40–200 ms~\cite{doane1980interactions, volkmann1980eyeblinks}, roughly 6 frames per blink are detected at 30fps, making blink removal crucial for data quality.

The final dataset~\footnote{\textbf{https://www.kaggle.com/datasets/vijuls/PupilDiameterDatasets}} includes 212,073 eye images filtered from 226,912 frames after preprocessing and blink removal. Left and right eye images, along with depth maps, are stored in directories, with a CSV file logging timestamps, session IDs, gaze, pupil data, and frame paths. Sample CSV files, cropped eye images, and depth maps are included in the supplementary materials.

\begin{table}[t!]
\caption{Leave one participant out cross validation (LOPOCV) of ResNet18 and ResNet50, evaluated separately for left and right eyes. We excluded one participant per training run and tested the model performance on the left-out participant. This process was repeated for all participants, with the table summarizing the mean and standard deviation of performance metrics across all runs.}
\scalebox{1.0}{
    \centering
    \renewcommand{\arraystretch}{1.0}
    \begin{tabular}{ccc}
    \toprule
    \textbf{Eye} & \textbf{Model} & \textbf{MAPE} $\downarrow$ \\ 
    \midrule
    Left    & ResNet18 & 3.411629\% $\pm$ 1.966436\% \\
            & ResNet50 & \textbf{3.234711\%} $\pm$ 2.032996\% \\
    \midrule
    Right   & ResNet18 & 4.288911\% $\pm$ 2.446597\% \\
            & ResNet50 & \textbf{3.644096\%} $\pm$ 1.769516\% \\
    \bottomrule
    \end{tabular}
}
\label{tab:cross_results}
\end{table}

\section{Model Training and Results}
We trained ResNet~\cite{he2016deep} models using leave-one-participant-out cross-validation (LOPOCV). ResNet18 and ResNet50 were trained for 50 epochs with a batch size of 128, using the Adam optimizer, 0.01 weight decay, and an initial learning rate of 0.001, reduced by 0.2 every 10 epochs. Mean Absolute Error (MAE) was used as the loss metric, and Mean Absolute Percentage Error (MAPE) quantified the results. ResNet50 consistently outperformed ResNet18, achieving lower MAPE for both left and right eyes. ResNet50 recorded a validation MAPE of 3.234711\% $\pm$ 2.032996\% for the left eye, compared to ResNet18’s 3.411629\% $\pm$ 1.966436\%. Similar trends were observed for the right eye, see \autoref{tab:cross_results}.

Instead of developing advanced deep learning models, our contribution and emphasis lies in providing publicly available dataset and the development of a practical web application for real-world pupil diameter estimation without specialized hardware. The best-performing ResNet models were integrated into our web application, \webAppName, and deployed on Hugging Face Spaces\footnote{\textbf{https://huggingface.co/spaces/vijulshah/pupilsense}}, enabling public access and advancing pupilometry research.

\section{Web Application: \webAppName}
We present a novel web application \webAppName shown in \autoref{fig:application}.
The application estimates pupil diameters from everyday images and videos. The application provides an in-depth analysis of the recordings, including Class Activation Maps (CAMs), which show the activated areas of the model influencing the output values based on the given input image, various graphs illustrating the predicted diameter values for the left and right pupils, Eye Aspect Ratios (EAR) in the event of blinking and a data frame table consolidating all the values in a single view for each frame in video inputs.

Within the application, users can upload photos or videos featuring either a person's entire face or a close-up of the left or right eye. These images or videos can be captured using a smartphone or a webcam. The application allows users to adjust several settings, for instance, on uploading the media, users can select which pupil diameters to estimate, choosing from two primary options: 

\begin{itemize}
    \item \textit{Both Pupils}: This option automatically detects both the left and right eyes in the uploaded media, utilizing separate models to estimate the pupil diameters for each eye. The application identifies the face and crops out the eyes accordingly. 
    \item \textit{Single Pupil (Left or Right)}: Users can focus exclusively on the left or right eye. The application applies the corresponding model to estimate the diameter of the chosen pupil.
\end{itemize}

Users also have the option to choose between two model architectures: \textit{ResNet18} or \textit{ResNet50}. 
Moreover, the application can detect blinks in the uploaded media if this feature is activated. This functionality employs an EAR threshold along with a \textit{Vision Transformer (ViT)} model, analogous to the blink detection method described in \autoref{sec:preprocessing}. After uploading the media and configuring the desired settings, users can click the \textit{Predict Diameter and Compute CAM} button to display results next to the uploaded media. 

For images, the results include: (1) a cropped view of the left, right, or both eyes, depending on the user’s selection, (2) a CAM illustrates the areas activated in the model’s last convolutional layer based on the input image, and (3) the estimated pupil diameter for each eye (or the selected pupil).

For videos, results are displayed frame by frame, with each frame showcasing: (1) the cropped eye image, (2) the corresponding CAM image, and (3) the estimated pupil diameter for that specific frame.

As soon as the video starts processing, the resulting frames are played in a continuous loop at ten frames per second (\autoref{fig:application} - B) for easy viewing. For uploaded videos, an interactive line chart (\autoref{fig:application} - C) illustrating the estimated pupil diameter for the left or right pupil (or both) appears below the results. If blink detection is activated, the frames corresponding to blinking will not display estimated pupil diameters, resulting in gaps in the graph for those frames, as seen in (\autoref{fig:application} - C). When blink detection is enabled, each frame's additional graph representing the EAR (\autoref{fig:application} - D) is displayed. This visual aid facilitates the identification of blinks, with marked horizontal green and red lines indicating the acceptable, rejectable, and gray zones for blink detection. Lastly, a data frame containing predicted diameters and EAR values is shown at the bottom in a Table format (\autoref{fig:application} - E), which can also be downloaded as a CSV file. It also contains two additional columns - one indicating the difference between the predicted diameter values for the corresponding frame and another indicating which one was greater. 

The development of this app, along with its ability to visualize data and predictions, allows end users to gain a detailed understanding of the dynamics of pupil diameters. By providing transparency into how the models perform on images and videos, users can see the prediction process in action. This openness helps build trust in the application, as users can observe the models used, predictions made, and the steps the app takes to generate those predictions.

\section{Limitations and Future Work}

\webAppName currently supports only post-analysis, lacking real-time capabilities due to resource-limited hosting. More efficient models suitable for low-resource environments should be integrated to overcome this. Additionally, implementing real-time processing and estimation in the background while performing tasks related to specific medical diagnoses, such as ADHD, Alzheimer’s, Parkinson’s, or schizophrenia, is a potential area for future development. Additionally, validating models with diverse cameras, leveraging depth maps for estimations, and collaborating with ophthalmologists to improve data collection are essential. Mobile phone cameras should also be explored as a source for data.

The dataset's reliance on a single camera model and exclusion of participants with eyeglasses or health conditions may impact its robustness. Additionally, the small size of pupil images limits feature extraction. Applying super-resolution (SR) techniques such as HAT~\cite{chen2023hat} or SRResNet~\cite{ledig2017photo} and fine-tuning these models on eye-cropped datasets like FFHQ~\cite{karras2018style} or CelebA-HQ~\cite{huang2018introvae} could enhance detail, as shown in \textit{Shah et al.}~\cite{shah2024webcam}. Combining SR with Pix2Net~\cite{jin2024pix2next} for RGB-to-NIR image translation may further improve accuracy, especially in low-contrast conditions where distinguishing pupil features in darker irises is challenging.

\section{Societal Impact}
Our web application, \webAppName, prioritizes user privacy by not storing personal data. The dataset is released under the CC BY-NC 4.0 license, encouraging ethical, non-commercial use. Images and pupil diameter data are anonymized with low-resolution, cropped eye regions to prevent personal identification.

Despite these measures, biases in the training data—such as the lack of certain nationalities or individuals wearing eyeglasses—may impact the fairness and accuracy of estimations for underrepresented groups.
Future research should aim to address these biases to enhance pupil diameter estimation systems, while also developing models that operate locally on users' devices or transmit only cropped eye data to server-hosted models. This would further protect user privacy, ensure data sovereignty, and minimize the risks associated with the misuse of facial data.

\section{Conclusion}
In this work, we introduced \webAppName, a web application, along with a collected dataset aimed at advancing eye monitoring research by enabling the development of models that estimate pupil diameter using standard webcam images. 
Our dataset significantly contributes by addressing the shortage of publicly available datasets that provide eye images paired with precise pupil diameter annotations. Focusing on recordings from the webcams, our dataset opens up opportunities for pupil-related research in low-resource environments and everyday computing contexts. Our results demonstrate that models trained on our dataset, particularly the ResNet50 architecture, perform well in estimating pupil diameters. Additionally, we show the practicality of a web-based application for pupil diameter estimation that goes beyond controlled lab environments, offering an accessible solution for users without specialized technical knowledge, making it usable in natural, everyday settings. 

\section*{Acknowledgements}
This work was supported by the DFG International Call on Artificial Intelligence ``Learning Cyclotron'' (442581111) and the BMBF project SustainML (Grant 101070408).

{\small
\bibliographystyle{ieee_fullname}
\bibliography{egbib}
}

\clearpage
\newpage
\section*{Supplementary Material}
\subsection*{Dataset Splits}
We employed a 5-fold cross-validation technique to split the dataset. The participants included in each validation and test fold are listed in the ~\autoref{tab:5_fold_cv}. For each fold, the remaining participants were used for training.

\begin{table}[h!]
\centering
\begin{tabular}{|c|c|c|}
\hline
\textbf{Fold} & \textbf{Validation Set} & \textbf{Test Set} \\
\hline
Fold-1 & [3, 7, 15, 44, 51] & [1, 4, 6, 25, 36] \\
\hline
Fold-2 & [4, 8, 16, 45, 50] & [2, 5, 7, 26, 37] \\
\hline
Fold-3 & [8, 12, 22, 34, 47] & [3, 16, 26, 38, 43] \\
\hline
Fold-4 & [5, 13, 23, 33, 41] & [9, 19, 29, 39, 49] \\
\hline
Fold-5 & [1, 11, 20, 32, 48] & [10, 14, 24, 28, 31] \\
\hline
\end{tabular}
\caption{5-Fold Cross-Validation Scheme detailing participants for each fold}
\label{tab:5_fold_cv}
\end{table}

\subsection*{Model Details}
We used ResNet18 and ResNet50 to train and evaluate our dataset. The models were originally designed for 224 x 224 dimension images. Given that our dataset images are 16 x 32, we upsampled them 2 times using bicubic interpolation to reach 32 x 64 dimensions. We then zero-padded the width and height to achieve 224 x 224 dimensions. Additionally, we incorporated a linear layer with a single output as a regression head for each model, to estimate the diameters of the left and right pupils separately.

\subsection*{Training Details}
Both ResNet18 and ResNet50 were trained from scratch for 50 epochs, separately for the left and right eyes, with a batch size of 128. We used the AdamW optimizer with default settings, a weight decay of 0.01, and an initial learning rate of 0.001. A learning rate scheduler decreased the learning rate by a factor of 0.2 every 10 epochs. We employed L1Loss (Mean Absolute Error) as the loss function.

\subsection*{Visualizations}
\begin{figure*}[t!]
  \centering
  \includegraphics[width=\linewidth]{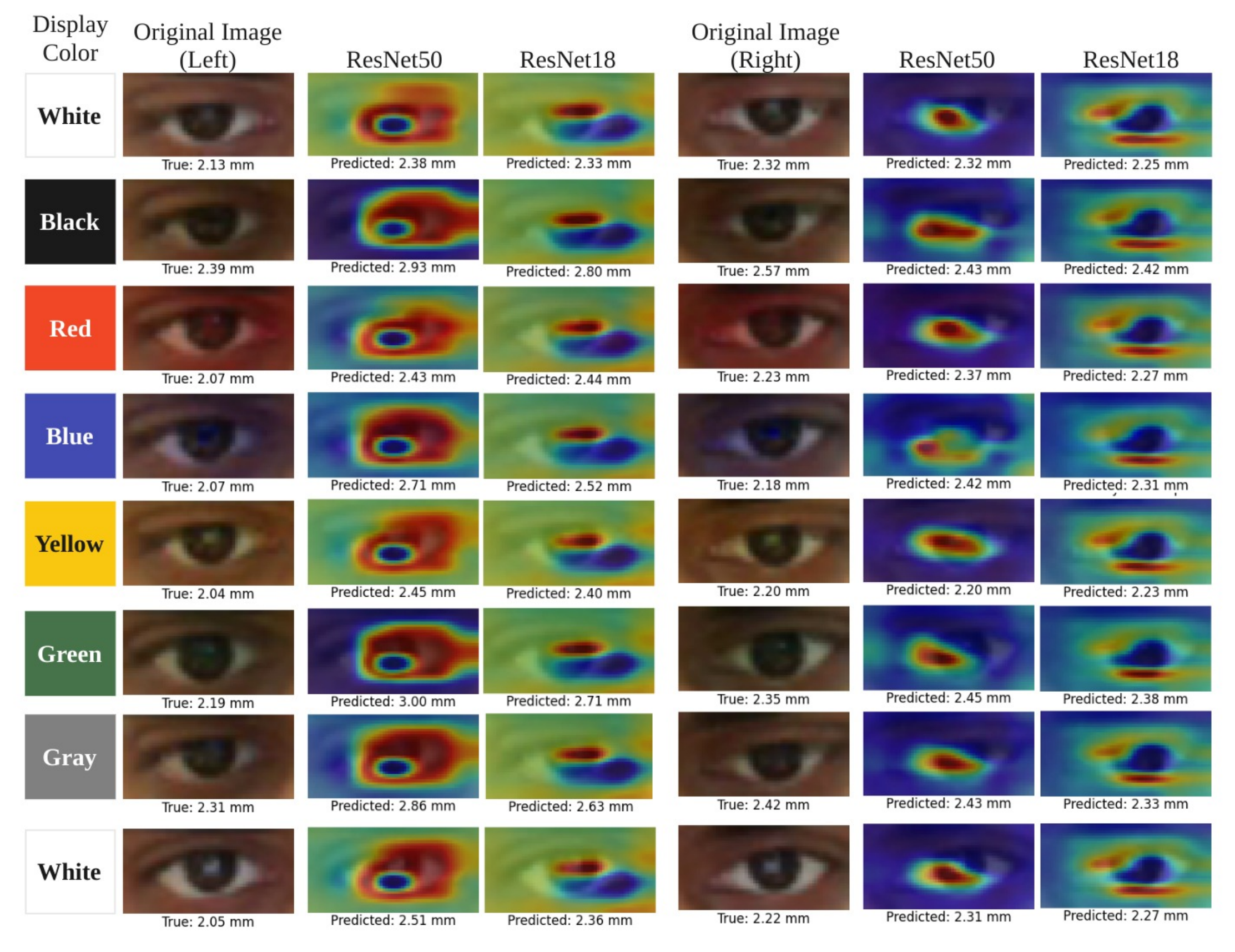}
  \caption{
   Class Activation Map (CAM) visualizations of ResNet50 and ResNet18 for a test participant's left and right eyes viewing different display colors on a monitor. True and Predicted values indicate the original and estimated pupil diameters of the left and right eyes in millimeters.
  }
  \label{fig:cam_vizs}
\end{figure*}

~\autoref{fig:cam_vizs} illustrates the Class Activation Map (CAM) of the last convolution layer of ResNet50 and ResNet18, evaluated on a test participant viewing different display colors. ResNet50 focuses on outer regions and color intensities for the left eye, while ResNet18 targets a small area within the iris. ResNet50 concentrates on the inner eye regions for the right eyes, whereas ResNet18 focuses on the surrounding edges and color variations. These results suggest that accurate pupil diameter estimation requires the model to pay attention to both the iris and the surrounding intensity changes.

\subsection*{Suggestions for Future Work}
\begin{figure}[t!]
    \centering
  \includegraphics[width=\linewidth]{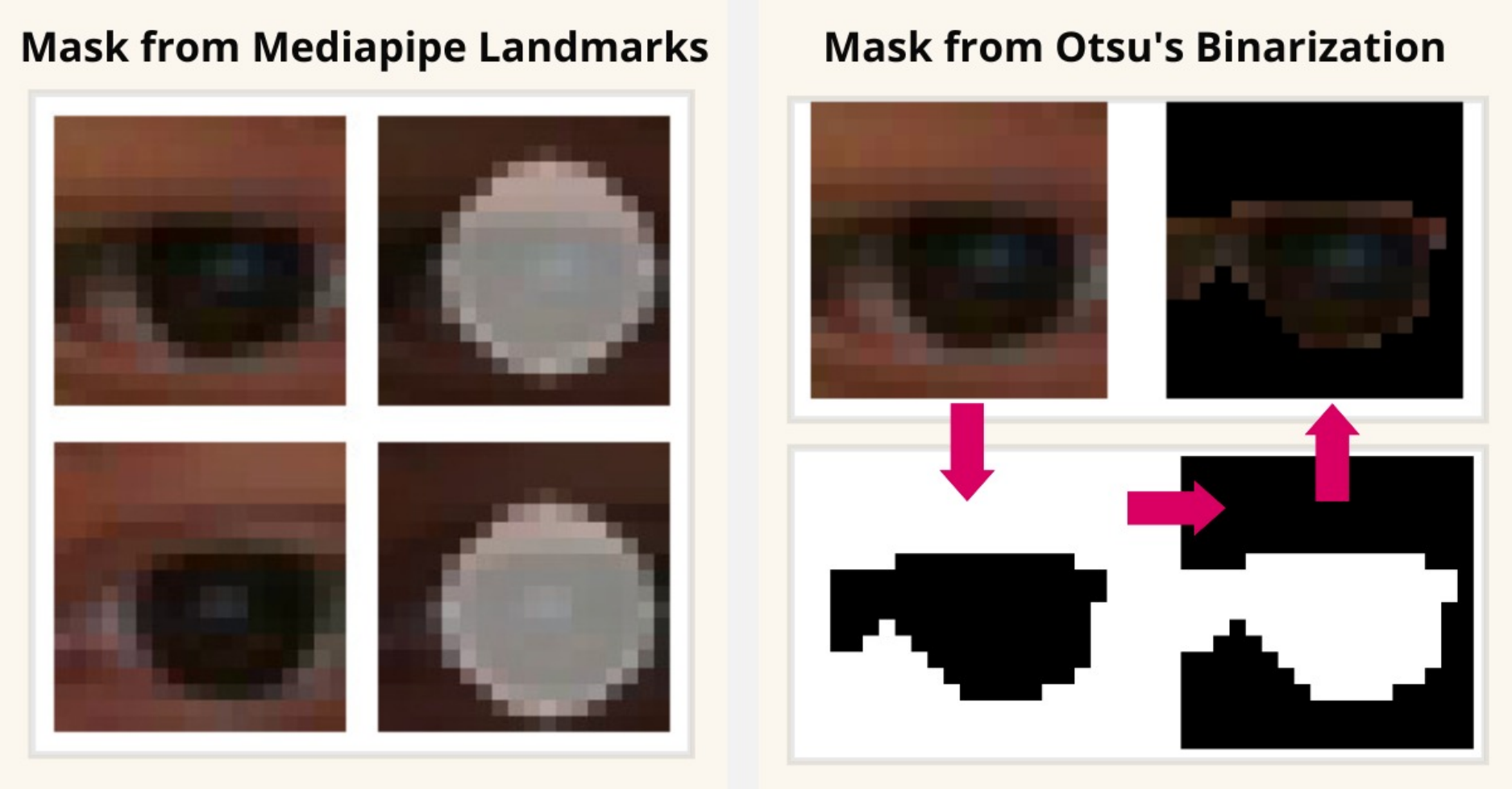}
  \caption{
    Iris masks were extracted using Mediapipe landmarks \textbf{(left)} and Otsu's Binarization \textbf{(right)}.
  }
  \label{fig:future_works}
\end{figure}

Mediapipe \cite{mediapipe_facelandmarker}, used for eye extraction, also provides landmarks of the detected iris region. These landmarks can be used to create a mask or apply image processing techniques like Otsu's binarization \cite{otsu_binarization}. These methods allow for the segmentation of the iris in the eye images, as shown in ~\autoref{fig:future_works}. These segmentation masks can be used in attention-based models to focus specifically on the iris region containing the pupil, potentially improving the accuracy of pupil size detection and gaze estimation.

\end{document}